\documentclass[journal,twoside,web]{ieeecolor}
\usepackage{generic}
\usepackage{cite}
\usepackage{amsmath,amssymb,amsfonts}
\usepackage{algorithmic}
\usepackage{graphicx}
\usepackage{textcomp}
\usepackage{siunitx}
\usepackage{url}
 \usepackage{relsize}
 \usepackage[caption]{subfig}
 \usepackage{pgfplots}
 \usepackage{tikz}
\usetikzlibrary{calc}

\def\BibTeX{{\rm B\kern-.05em{\sc i\kern-.025em b}\kern-.08em
    T\kern-.1667em\lower.7ex\hbox{E}\kern-.125emX}}
\newcommand*{\Rn}[1]{\ensuremath{\mathbb{R}^{n_{#1}}}}
\newcommand*{\R}{\ensuremath{\mathbb{R}}}

\newcommand\mydots{\hbox to 2mm{.\hss.\hss.}}

\newcommand*{\rev}[1]{\textcolor{black}{#1}}

\newtheorem{task}{Task}
\newtheorem{theorem}{Theorem}
\newtheorem{corollary}{Corollary}
\newtheorem{assumption}{Assumption}

\newtheorem{lemma}{Lemma}
\newtheorem{observation}{Observation}

\usepackage{fancyhdr}
\begin{document}
\title{Safe Machine-Learning-supported Model Predictive Force and Motion Control in Robotics}
\author{Janine Matschek, Johanna Bethge, and Rolf Findeisen
\thanks{Janine Matschek and Rolf Findeisen are with the Control and Cyber-Physical Systems Laboratory,  TU  Darmstadt, ({rolf.findeisen, janine.matschek}@iat.tu-darmstadt.de); 
Johanna Bethge ({johanna.bethge}@ovgu.de) is  with the Laboratory of Systems Theory and Automatic Control, Otto-von-Guericke-Universität Magdeburg. \\ \\
}
}

\maketitle

\begin{abstract}
Many robotic tasks, such as human-robot interactions or the handling of fragile objects, require tight control and limitation of appearing forces and moments alongside sensible motion control \rev{to achieve safe yet high-performance operation}. We propose a learning-supported model predictive force and motion control scheme that provides \rev{stochastic} safety guarantees while adapting to changing situations. Gaussian processes are used to learn the uncertain relations that map the robot's states to the forces and moments. The model predictive controller uses these Gaussian process models to achieve precise motion and force control \rev{under stochastic constraint satisfaction}. 
\rev{As the uncertainty only occurs in the static model parts - the output equations - a computationally efficient stochastic MPC formulation is used. Analysis of} recursive feasibility of the optimal control problem and convergence of the closed loop system \rev{for the static uncertainty case are given}. \rev{Chance constraint formulation and back-offs} are constructed based on the variance of the Gaussian process to guarantee safe operation. The approach is illustrated on a lightweight robot \rev{in simulations and experiments}. 
\end{abstract}



\section{Introduction}
\label{sec:introduction}
\IEEEPARstart{R}{obots} are increasingly used for interactive and cooperative tasks in a wide range of applications, which require safe interaction with the environment. 
For instance, robots should safely support humans in production processes, without the traditional separation between the robotic and human coworkers \cite{matheson2019human}. Robots, furthermore, should reliably assist physicians in  medical treatments\cite{matschek2020learning} , should securely support elderly people \cite{bedaf2015overview}, or should help in rehabilitation tasks \cite{maciejasz2014survey}. Safe interaction requires tight motion control limiting the forces and appearing moments.

While many force control schemes exist,  most do not ensure satisfaction of constraints, e.g., with respect to the appearing forces and moments \cite{villani2018survey, zacharaki2020safety, villani2016force}. Existing approaches are furthermore often limited with respect to the flexibility to formulate the force-motion control task in a structured manner.  

 This work proposes a flexible learning-supported model predictive motion and force control scheme, which ensures the safe satisfaction of force and motion constraints despite uncertainty in the wrench model. 

Model predictive control (MPC), an optimization based control strategy, allows  consider\rev{ing} constraints or limitations while optimizing the systems predicted behavior. 
MPC schemes are by now widely used for robotic tasks, including the explicit consideration or limitation of forces, see e.g., \cite{killpack2016model, muller2019model,kocer2019model, mitsioni2019data,gold2020iros}. 
A series of motion and force control schemes using MPC exist  \cite{matschek2017force,kazim2018combined, wahrburg2016mpc,bednarczyk2020model,gold2020model}, spanning from hybrid motion  and force control \cite{matschek2017force}, admittance control \cite{kazim2018combined, wahrburg2016mpc} to  unified approaches for MPC force control \cite{bednarczyk2020model,gold2020model}. 

Most results, however, rely on the availability of good force models. 
They can, e.g., be obtained via  the identification of tailored linear and nonlinear spring and damper models \cite{flores2016contact}. 
However, the resulting models 
are often limited in their precision. 
Machine learning approaches can be used to overcome this challenge, allowing for data-driven or hybrid modeling. 
We use Gaussian processes (GPs) \cite{kocijan2016modelling,rasmussen2006gaussian}  to capture and learn the state-force interactions.  Force modelling via Gaussian processes is, e.g.,  considered in \cite{vina2013predicting, li2017gaussian, horii2014contact, medina2019impedance,matschek2020direct}.
 Learning of friction and grasping forces using GPs is considered in \cite{vina2013predicting, li2017gaussian}, while \cite{horii2014contact} uses GPs to process sensor data from a tactile array.  Contact forces occurring in human-robot interactions are considered in \cite{medina2019impedance} via GPs. The  interaction of a robot with a static environment using GPs  to obtain a learned contact force model for MPC is considered in \cite{matschek2020direct}.    

This work proposes a combined robot motion and force control scheme. An  MPC scheme is presented assuming that forces and moments are described by an (unknown or uncertain) static mapping of the robot's states and inputs. This mapping is learned using GPs, while ensuring constraint satisfaction on the forces \rev{with high probability} to ensure safe interaction. Compared to  
\cite{killpack2016model, muller2019model,kocer2019model, darivianakis2014hybrid, nadeau2011automatic,lenz2015deepmpc, mitsioni2019data, erickson2018deep,sheng2004model, ladoiye2018control,gold2020iros,matschek2017force,kazim2018combined, wahrburg2016mpc, bednarczyk2020model,gold2020model, matschek2020direct},
 we propose to combine first-principle models, e.g., linear spring models, with GP models. 
The resulting hybrid model is used in a model predictive control scheme for combined motion  and force control. \rev{As only the static mapping from the states to the forces is learned, it is not necessary to propagate the influence of the resulting (``output") uncertainty through the dynamics to ensure safe operation, as often considered in other learning-supported MPC schemes \cite{ kocijan2005nonlinear, likar2007predictive, umlauft2018scenario,  hewing2019cautious, yang2015fault}}. 
Inspired by \cite{wang2015robust,soloperto2018learning, wang2016stochastic, hewing2019cautious, likar2007predictive,ostafew2016robust, grancharova2008explicit}, we utilize the variance in the hybrid output model to tighten the MPC constraints. In contrast to \rev{existing} works, however, we tighten only the constraints mapping from the states to the  \rev{``output" (=forces)}, instead of using  tubes \cite{langson2004robust} for state constraint tightening of the dynamics. 
\rev{Since the tightening is based on the variances it leads to a chance constraint formulation of the MPC. }

\begin{figure}[t]
\centering
\includegraphics[width=0.9\columnwidth,trim={0 0 0 4cm}, clip]{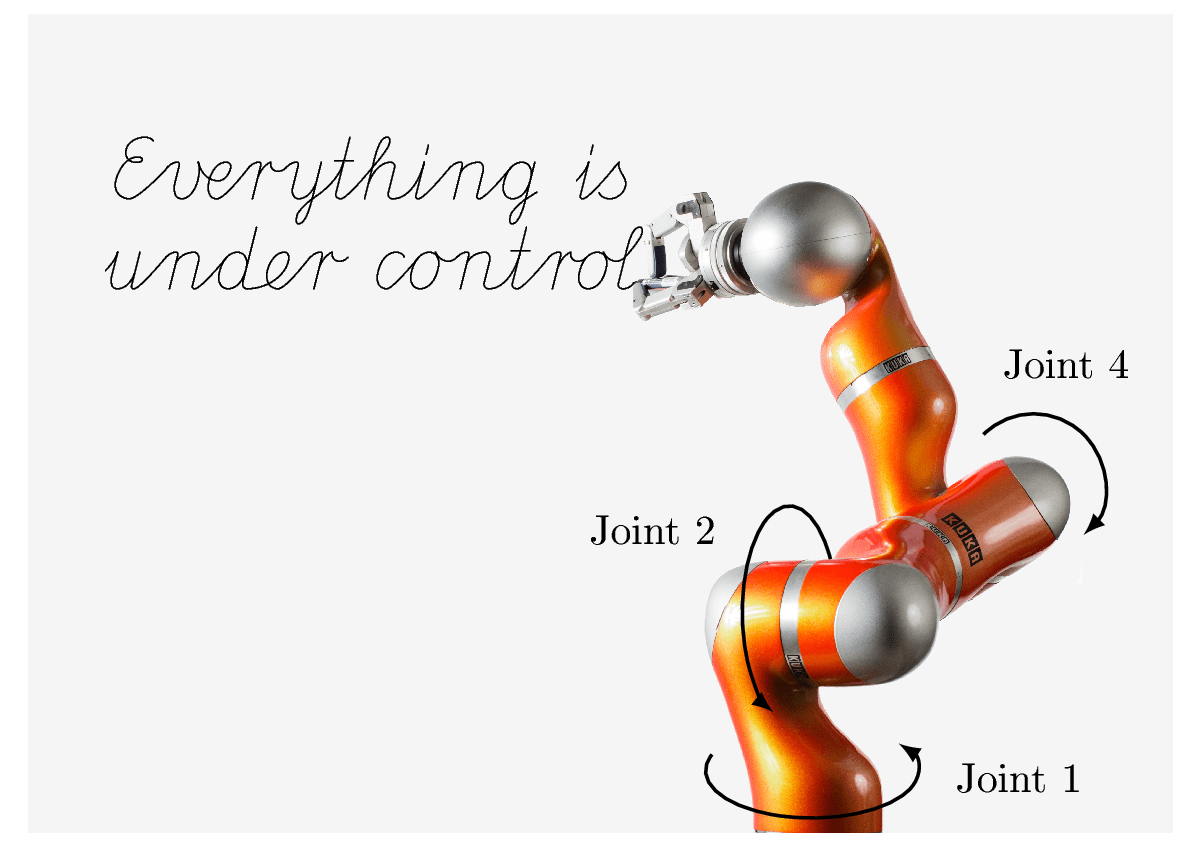}
\caption{\rev{Example application: a lightweight robot equipped with a robotic hand should write on a soft surface.}}
\label{fig:robo_writing}
\end{figure}

The main contribution of this work is the development of a safe force and motion control scheme combining Gaussian process model learning and MPC \rev{where model uncertainty enters the} static mapping from the states to the \rev{controlled variables}. 
We derive stability/recursive feasibility conditions for the  learning-supported predictive controller. \rev{In the predictive controller, the uncertainty estimates provided by the Gaussian process are integrated as chance constraints to guarantee stochastic constraint satisfaction. As the uncertainty is related to the ``static output" mapping, one can calculate it offline efficiently, i.e., there is no need for computational intense online propagation through the system dynamics.} The approach allows offline learning, as well as  learning in between batches of iterative tasks. Summarizing, the uncertainty estimates allow \rev{us} to find  suitable back-offs of the constraints to robustify the model predictive controller. 
The theoretical findings are illustrated in simulations and experiments considering  \rev{force and motion control of} a lightweight KUKA robot, see Figure~\ref{fig:robo_writing}. 

The remainder of the paper is structured as follows:
The \rev{considered, learning supported} control problem is introduced in Section~II. 
Section~III proposes the model predictive force controller. It is extended by a Gaussian-process-learning component, which is introduced in Section IV. The \rev{overall} resulting learning-supported force controller is presented in Section~V. 
 Section~VI illustrates the proposed approach \rev{in} simulations and experiments \rev{for a} lightweight robot. Finally, Section~VII concludes this work with a summary and an outlook.


\section{Problem Setup}\label{sec:task}

We aim to develop a control scheme that guarantees safe interaction and satisfaction of force constraints, e.g., between a robot and its environment. An example might be a robot that is supposed to write on a delicate surface limiting the maximum force, see Fig.~\ref{fig:robo_writing}, a robot that should polish a surface \rev{with desired intensity, or a robot that should interact safely with a human not exceeding a certain force.} 
To achieve this goal, we \rev{fuse  model-based predictive motion and force controller with Gaussian process based machine-learning to capture the occurring, often difficult to model, forces.}

MPC allows tackl\rev{ing} a wide variety of force and motion control tasks \cite{wahrburg2016mpc, matschek2017force, kazim2018combined, erickson2018deep, matschek2020direct, gold2020iros, gold2020model}, spanning from direct and indirect force control to joint  hybrid motion/position and force control. 
We focus on hybrid position and force control, where some positions, as well as specific forces/moments, should be controlled jointly\cite{siciliano2012robot, villani2016force}. 

Specifically, we focus on motions where a robot should move (with a tool) along a "surface", cf.  Fig.~\ref{fig:robo_writing}. To do so, the interaction forces need to be tightly controlled and should be limited for safety reasons. 
To account for changing environments and to improve performance, the controller should be able to adapt and learn based on data.
We formalize this task in the following sections.

\subsection{Constrained safe force and motion control}
We consider that the dynamical system for the motion and force control task, e.g., a robot and its environment, is described by the nonlinear dynamics
\begin{subequations}
\label{eq:nonlinear-system}
\begin{eqnarray}
\dot x(t) &=&  f( x(t), u(t)), \quad x(0) = x_0, \label{eq:nonlinear-system_stateequation}\\
y(t) &=& h( x(t) ). \label{eq:nonlinear-system_output}
\end{eqnarray}
\end{subequations}
Here, $t\in \R$ represents the time, $x(t)\in \Rn{\text{x}}$ denotes the state, and $u(t)\in \Rn{\text{u}}$ is the input of the system. 
The states might for example describe the angular positions and velocities of the robot, while the inputs might represent the torques in the joints. The map $f:\Rn{\text{x}} \times \Rn{\text{u}} \to \Rn{\text{x}}$ describes the dynamics, while the ``output" $y(t)\in \Rn{\text{y}}$, given by $h:\Rn{\text{x}} \to \Rn{\text{y}}$ 
maps the states to the variables of interest -- the controlled variables, such as  positions, velocities, forces and moments\footnote{\rev{The output $y$ might also explicitly depend on the input $u$. While the derived results hold for this case, we do not consider the influence of the inputs directly for simplicity of presentation.}}.  \rev{We assume, for simplicity of presentation, that full state measurements are availabe. Note that the output does not relate to measured variables, it is used to evaluate the control performance via the controlled variables. Section \ref{sec:example} presents an example for force and motion control of a lightweight robot, cf.  Fig. \ref{fig:robo_writing}. } 

\rev{We want to satisfy constraints on the states, inputs, and outputs, which are given by the sets $\mathcal{X}\subset\Rn{\text{x}}$, $\mathcal{Y}\subset\Rn{\text{y}}$, and $\mathcal{U}\subset\Rn{\text{u}}$, i.e., to limit the forces, torques, or states. Depending on the formulation, these will be hard -- set based -- constraints, or probabilistic constraints, see Section~\ref{Chapter:learning_MPC}. With respect to the sets we assume that
\begin{assumption}
The sets $\mathcal{X}$, $\mathcal{Y}$ are closed, and  $\mathcal{U}$ is compact.\label{ass:compact_sets_xu}
\end{assumption}
With respect to the dynamical system we furthermore assume that:
\begin{assumption} 
$f$, and $h$ are  sufficiently often continuously differentiable and locally Lipschitz\footnote{\rev{One might relax this condition to only hold in a region of interest, which we avoid to streamline the presentation.}}.
\rev{Furthermore, for any continuous input signal with $u(\tau)\in\mathcal{U}$ and for all $x_0\in \mathcal{X}$, \eqref{eq:nonlinear-system} admits a unique absolutely continuous solution.}
\label{ass:Lipschitz_xy_unique}
\end{assumption}}
The goal is to follow a given reference \rev{path}, which is defined as a geometric curve in the ``output" space, as good as possible while satisfying constraints. For example the robot shown in Fig.~\ref{fig:robo_writing} should write on a flexible and possibly uneven surface. Hence, it should follow a given path while keeping  forces at desired values and constrained, \rev{even under disturbances and uncertainty}. The reference path is given by  
\begin{equation}
\label{eq:pf:path}
\mathcal{P} := \left\{ y_\text{r} \in \mathbb{R}^{n_\text{y}} \vert y_\text{r} = r_\text{pf}( \theta(t) ) \right\},
\end{equation}
with the parametrization $r_\text{pf}: \Theta \to \mathbb{R}^{n_\text{y}}$.
Note that the reference only indirectly depends on the time via the path parameter $\theta(t) \in \Theta=[-1,0]$.  The evolution of the path parameter over time is not fixed a priori -- it is adjusted by the controller online \cite{matschek2019nonlinear,faulwasser2016nonlinear,matschek2017force}, which provides additional degrees of freedom that can be exploited by the controller. For example, in case of disturbances, the controller can adjust the reference speed while compensating for the disturbance, thus avoiding performance deterioration, cf. \cite{matschek2017force}. 

Given the  reference path $\mathcal{P} $, we aim to solve the following task  \cite{faulwasser2009nonlinear,faulwasser2016nonlinear,matschek2017force}:

\begin{task}[Path Following Force  and Motion Control]
Given the system~\eqref{eq:nonlinear-system} and path~$\mathcal P $, design a controller that achieves:
\begin{enumerate}
\item Convergence: The ``output" \eqref{eq:nonlinear-system_output} should converge to the set $\mathcal{P}$, i.e., 
$\lim\limits_{t \to \infty} y(t)-r_\textup{pf}(\theta(t)) = 0$.
\label{subtask:convergence_pf}
\item Forward Motion along the Path: The reference $r_\textup{pf}(\theta(t))$ moves along
$\mathcal{P}$ in the direction of increasing $\theta$ values, i.e., $\dot \theta(t) \geq 0$ and $\lim\limits_{t\to \infty} \theta(t)=0$.
\label{subtask:forward_motion}
\item Constraint Satisfaction: The states, inputs, and ``outputs" should satisfy   $x\!\in\! \mathcal{X}$,  $u\! \in\! \mathcal{U}$, and $y\! \in\! \mathcal{Y}$. \label{subtask:constraints_pf}
\end{enumerate}
\label{task:path_following}
\end{task}
Note that $y$ and the path $\mathcal{P}$ can describe  (desired) motions and forces/moments. Hence, one can include constraints on the forces by choosing appropriate constraint sets.

\rev{In comparison to earlier works \cite{faulwasser2009nonlinear,faulwasser2016nonlinear,matschek2017force},  we consider that the ``output"  mapping $h$ might be unknown or uncertain, as relations  between the robot's states, the environment and the forces/moments  are, in general,  challenging  to model. To do so, we use methods from machine learning, as outlined in the following subsection.} 

\subsection{Learning-supported force/output modeling}
We propose to use a hybrid model combining machine learning and first principle model components. The hybrid model is used in the controller to predict, control, and limit the forces for a reliable and safe operation. 
It is given by 
\begin{equation}
    \tilde h(x)=h_\mathrm{fp}(x)+h_\mathrm{ml}(x),
    \label{eq:hybrid_output}
\end{equation}
 where the first principle part $h_\text{fp}:\Rn{\text{x}} \to \Rn{\text{y}}$ encodes prior knowledge, while $h_\text{ml}:\Rn{\text{x}} \to \Rn{\text{y}}$ is obtained from data via machine-learning \rev{achieving the following task:}




\begin{task}[Output/Force Model Learning]
Learn a hybrid model $\tilde h(x)=h_\mathrm{fp}(x)+h_\mathrm{ml}(x)$ exploiting  data $D$  such that:
\begin{enumerate}

\item  $\tilde h(x)$ fits the data, i.e., $\tilde h(x_i) \approx \hat y_i$ for $(x_i,\hat y_i)\in D$, while trading off between complexity and consistency. 
\item The error $(\tilde h(x)-h(x)) $ can be bounded.

\end{enumerate}
\label{task:GP}
\end{task}
We outline in Section~\ref{sec:GP} how to use Gaussian processes to achieve this task.  We use the resulting hybrid model in a tailored model predictive control scheme to achieve safe motion and force control, as outlined in Section~\ref{Chapter:learning_MPC}.


\section{Model Predictive Force and Motion  Control}\label{sec:MPC_setup}
Model predictive control is a model based approach that repeatedly solves an optimal control problem \cite{rawlings2017MPC,findeisen2003state}. 
It allows control\rev{ling} linear and nonlinear systems, \rev{the} direct consideration of constraints, as well as preview information about disturbances, changing references, and paths \cite{matschek2019nonlinear}. 
The possibility to explicitly account for constraints makes \rev{MPC a valuable tool for the control and decision-making of autonomous systems and robots. Constraint consideration enables safe operation by limiting torques, robot positions, velocities, or forces and taking obstacles directly into account, as outlined in the following.}

\subsection{Optimal control problem formulation}
\rev{For path following motion and force control (Task~\ref{task:path_following}) we propose to use a tailored sampled-data MPC formulation  \cite{faulwasser2009nonlinear,faulwasser2016nonlinear,matschek2017force}.} At every sampling time  $t_k$ an  optimal control problem, exploiting the reference speed along the path as an additional degree of freedom, is solved:
\begin{subequations} \label{eq:OCP_pf}
\begin{equation}
 \underset{\bar u,\bar v}{\text{min}} ~ J_\text{pf}\left(\bar e_\text{pf},\bar \theta,\bar u,\bar v,\bar x,\bar z \right) \label{eq:min}
\end{equation}
subject to $\forall \tau\in[0,T]$
\begin{align}
     \dot{\bar x}(\tau)& = f(\bar  x(\tau),\bar  u(\tau)), \quad \bar  x(0) = x(t_k), \label{eq:con_x_pf}\\
     \dot{\bar z}(\tau)& = g(\bar  z(\tau),\bar  v(\tau)), \quad \;\bar  z(0) = z(t_k), \label{eq:con_z_pf}\\
 \quad \bar  \theta (\tau) &=   l(\bar  z(\tau)), \label{eq:con_theta_pf} \\
 \quad \bar  e_\text{pf}(\tau) &= r_\text{pf}(l(\bar  z(\tau))-h(\bar  x(\tau))  , \label{eq:con_e_pf} \\
 \quad \bar  x(\tau) &\in \mathcal{X}, ~ \bar u(\tau) \in \mathcal{U},~h(\bar  x(\tau)) \in \mathcal{Y}, \label{eq:con_X_pf} \\
 \quad \bar  z(\tau) &\in \mathcal{Z}, ~\bar  v(\tau) \in \mathcal{V},~\bar \theta(\tau) \in \Theta, \label{eq:con_Z_pf} \\
	(\bar  x(T&), \bar z(T))^\top \in \mathcal{F}_\text{pf}\label{eq:term_con_pf}
	\end{align}
\end{subequations}
Predictions are indicated by $\bar \cdot$. Besides the system dynamics \eqref{eq:con_x_pf},  a ``virtual reference  dynamics" \eqref{eq:con_z_pf}-\eqref{eq:con_theta_pf} is used, which allows shap\rev{ing} the dynamics of the speed along the path \cite{faulwasser2016nonlinear,matschek2017force}.  The virtual (reference) state, input, and output are $z(t)\in \Rn{\text{z}}$, $v(t)\in \R$, and $\theta(t)\in \R$, which are described by the  virtual system dynamics $g: \Rn{\text{z}}\times \R \to \Rn{\text{z}}$ and the virtual output $l:\Rn{\text{z}}\to \R$. The virtual input $v$ provides an additional degree of freedom in the optimal control problem allowing to  adjust the reference path progress/evolution.
\rev{The constraint \eqref{eq:term_con_pf} $\mathcal{F}_\text{pf} \subseteq (\mathcal{X} \times \mathcal{Z})\cap (h^{-1}(\mathcal Y) \times l^{-1}(\Theta))$ is a final terminal region constraint. It is used to ensure repeated feasibility and stability/convergence \cite{faulwasser2016nonlinear,matschek2017force}.}

\rev{The cost function penalizes the error $e_\text{pf}(t)$ \eqref{eq:con_e_pf} to achieve path following:}
\begin{align}
 \label{eq:J_pf}
 J_\text{pf}\left(\bar e_\text{pf},\bar \theta,\bar u,\bar v,\bar x,\bar z \right)
    := \int\limits_{0}^{T}&
      L_\text{pf}\left( \bar  e_\text{pf}(\tau), \bar  \theta(\tau), \bar u(\tau),\bar  v(\tau) \right)\text{d}\tau \nonumber\\
      + &E_\text{pf}\left( \bar  x(T), \bar z(T) \right).
\end{align}
Here, $L_\text{pf}: \Rn{\text{y}} \times \R \times \Rn{\text{u}} \times \R \to \mathbb{R}_0^+$ is a stage cost and $E_\text{pf}: \Rn{\text{x}} \times \Rn{\text{z}} \to \mathbb{R}_0^+$ is a terminal penalty term. 

State, input and output constraints are enforced by \eqref{eq:con_X_pf}, \eqref{eq:con_Z_pf} (Task \ref{task:path_following}.\ref{subtask:constraints_pf}). Forward motion along the path (Task \ref{task:path_following}.\ref{subtask:forward_motion})  is ensured  by  requiring that  $\Theta:=[-1, 0]$ and  that $\dot \theta\geq0$. 

\rev{The optimal control problem is solved at all sampling times $t_k$. From the resulting optimal input signal, only the first part until the next sampling instant is used and the optimization is repeated in a receding-horizon fashion.}

\rev{Convergence to the path/stability of path following MPC can be guaranteed, similar to standard MPC, by suitable choice of the cost function and the terminal constraints
 \cite{faulwasser2016nonlinear,matschek2019nonlinear}. \rev{ We consider that the following assumptions and conditions hold: }
\begin{assumption}
\label{ass:pf1}
The stage cost $L_\textup{pf}: \mathbb{R}^{n_\textup{y}} \times\mathbb{R}\times \Rn{\textup{u}} \times \Rn{\textup{v}} \to \mathbb{R}_0^+$ is continuous 
and lower bounded by a class $\mathcal{K}_\infty$ function $\alpha_1$ such that
$L_\textup{pf}(e_\textup{pf},\theta,u,v) \geq \alpha_1(\|(e_\textup{pf}, \theta-\theta_\textup{end})^\top \|)$.
\end{assumption}
\begin{assumption}
\label{ass:pf4}
The terminal cost $E_\textup{pf} : \Rn{\textup{x}} \times \Rn{\textup{z}} \to \R^+_0$ is positive semi-definite and cont. differentiable in $x$ and $z$, and the terminal set $\mathcal{F}_\textup{pf} \subseteq \mathcal{X} \times \mathcal{Z}$ is closed. Furthermore, for all $(\tilde x,\tilde z)^\top \in \mathcal{F}_\textup{pf}$ there exist inputs $(u_\mathcal{F},v_\mathcal{F})^\top (\cdot) \in \mathcal{U \times V}$ such that for all $\tau \in [0, T_\textup{s})$
\begin{align*}
\left(\begin{smallmatrix}
\dfrac{\partial E_\textup{pf}}{\partial x}, & \dfrac{\partial E_\textup{pf}}{\partial z}
\end{smallmatrix} \right)
\cdot
\left(\begin{smallmatrix}
f(x(\tau),u_\mathcal{F}(\tau))\\
g(z(\tau),v_\mathcal{F}(\tau))
\end{smallmatrix} \right) \\
+ L_\textup{pf}(e_\textup{pf}(\tau), \theta(\tau), u_\mathcal{F}(\tau), v_\mathcal{F}(\tau))\leq 0
\end{align*}
and that $x(\tau)= x(\tau,\tilde x|u_\mathcal{F})$ and $z(\tau)= z(\tau,\tilde z|v_\mathcal{F}) $ stay in $\mathcal{F}_\textup{pf}$, i.e., $\mathcal{F}_\textup{pf}$ is control invariant.
\end{assumption}}

Following \cite{faulwasser2016nonlinear,matschek2019nonlinear} one can establish the following result:
\begin{theorem}
\label{theo:pf} 
If Assumptions \ref{ass:compact_sets_xu}-\ref{ass:pf4} hold, and if the optimal control problem \eqref{eq:OCP_pf} is initially feasible, then \eqref{eq:OCP_pf} is recursively feasible and the path-following error $e_\textup{pf}$ converges to zero under sampled-data NMPC.
\end{theorem}
For details we refer to \cite{faulwasser2016nonlinear,matschek2019nonlinear}. We  use the outlined path following MPC scheme as a basis and adjust it in Section \ref{Chapter:learning_MPC} to guarantee constraint satisfaction/safety despite the uncertainty due to the hybrid uncertain output model.


\section{Gaussian processes}
\label{sec:GP}

We propose to use Gaussian processes to learn the unknown output mapping $h_\text{ml}$.  
Gaussian processes (GPs) are stochastic processes that follow an infinite-dimensional joint normal distribution \cite{kocijan2016modelling, rasmussen2006gaussian}. 
They have gained  increasing attention in the control community  \cite{yang2015risk, hewing2020on, kocijan2003predictive, cao2017gaussian, klenske2016gaussian, hewing2018cautious, hewing2019cautious}. 
We propose to use GPs, as they provide robustness against noise, inclusion of physical knowledge, and their stochastic nature allows fit\rev{ting} data without overfitting (Tasks~\ref{task:GP}.1 and \ref{task:GP}.2). 
Furthermore, GPs provide a posterior uncertainty measure to calculate a reliability bound (Task~\ref{task:GP}.3), which can be used to achieve safe operation as outlined in Section~\ref{sec:robustCase}.

\subsection{Gaussian process force modeling}

In the following we focus on modeling  a part of the output\rev{/controlled variable}, e.g., the contact forces, in dependence of the state  of the system/robot. 
In general, multiple output parts  $F\in \R^{n_F}$ with $n_F>1$ can be considered. In such a case multiple one-dimensional GPs or a multidimensional GP could be used. 
 For dimension $i$, we assume that noisy observations $\hat F_{i}$ are available that origin from 
 \begin{equation}
\hat F_{i}=h_{\text{i}}(x)+\eta_i
\label{eq:noisy_model}
\end{equation}
where the mapping $h_{i}$ \rev{(the considered force component of $h$)} from the system state $x \in \R^{n_x}$ to the measured variable $\hat F_{i} \in \R$ is corrupted by the uncertainty or noise $\eta_i$. 
This noise is assumed to follow an independent, identically distributed Gaussian distribution $\eta_i \sim \mathcal{N}(0,\sigma_i^2)$ with zero mean and variance $\sigma_i^2$.

We use a GP to model the underlying
function $h_{\text{ml},i}$ \rev{in $\tilde h_i$ which approximates   $h_i$} via
\begin{equation}
h_{\text{ml},i}(x) \sim \mathcal{GP} \left(m(x),\kappa\left(x,x'\right)\right).
\label{eq:GP}
\end{equation} 
Here,  $m(x)\in \R$ denotes the mean of the GP and $\kappa(x,x')\in \R$ denotes the covariance. This covariance is a measure of the joint variability of two random variables and the function $\kappa:\Rn{x}\times\Rn{x}\to \R$  is positive semi-definite  and symmetric. Once a prior assumption on the mean and covariance function is posed, the training of the GP is performed to obtain posterior  distributions. 
The GP is trained on the data $D:=(\boldsymbol{x} ,\boldsymbol{\hat F})$, where $\boldsymbol{x}:=(x^{1},\ldots, x^{n_D})^\top$ and $\boldsymbol{\hat F}:=(\hat F^{1}\rev{-h_{\text{fp},i}(x^1)},\ldots, \hat F^{n_D} \rev{-h_{\text{fp},i}(x^{n_D})})^\top$. The superscript $i$ in $x^i, \hat F^i_\text{n}$ with $i=1,2,\ldots,n_D$ enumerates the available measuring instances. 
The training determines the hyperparameters $\phi \in \Rn{\phi}$ of the GP, which include the parameters of the mean and covariance functions as well as the noise variance. The number of the hyperparameters $n_\phi$ depends hereby on the assumed prior structure, i.e., on the specific choice of the mean and covariance functions. 
Often an estimate of the most likely hyperparameters is obtained via the maximization of the logarithmic marginal likelihood \cite{rasmussen2006gaussian,kocijan2016modelling}.  
Given these hyperparameters, the posterior distribution of the GP conditioned on the prior and the data can be derived. 
The joint posterior distribution at previously seen data points $\boldsymbol{x}$ and at the query point $x^*$ is given by 
 
\begin{equation*}
\begin{pmatrix}
\boldsymbol{\hat F}\\ h_{\text{ml},i}(x^*)
\end{pmatrix} \sim \mathcal{GP}\left(   \begin{pmatrix}
\boldsymbol{m}(\boldsymbol{x})\\m(x^*)
\end{pmatrix}  , \begin{pmatrix}
K+\sigma^2 I & {k}\\ {k}^\top & \kappa(x^*,x^*)
\end{pmatrix}\right),
\end{equation*}
where $\boldsymbol{m}(\boldsymbol{x}):=(m(x^1), \ldots,m(x^{n_D}) )^\top$.
The covariance matrix $K$ specifies the covariance between all of the training data points and is given by 
\begin{equation*}K=\begin{pmatrix}
\kappa(x^1,x^1) & \cdots & \kappa(x^1,x^{n_D})\\
\kappa(x^2,x^1) & \cdots & \kappa(x^2,x^{n_D})\\
\vdots& \ddots& \vdots\\
\kappa(x^{n_D},x^1) & \cdots & \kappa(x^{n_D},x^{n_D})
\end{pmatrix}.
\end{equation*}
 The cross covariance between the trainings and the query points is given by $k:=(\kappa(x^1,x^*), \kappa(x^2,x^*),\ldots, \kappa(x^{n_D},x^*))^\top$.
Hence, the GP posterior is given by the posterior mean function $m^+:\mathbb{R}^{n_x} \to \mathbb{R}$ that is defined by
\begin{align}
m^+(x^*):=& m(x^*)
+{k}^\top (K+\sigma^2 I)^{-1}\big(\boldsymbol{\hat F}-\boldsymbol{m}(\boldsymbol{x})\big)
\label{eq:mean}
\end{align}
 and the posterior covariance $\kappa^+: \mathbb{R}^{n_x}\times \mathbb{R}^{n_x} \to \mathbb{R}$ that is given by
 \begin{align*}
 \kappa^+(x^*,x^*):=&\kappa(x^*,x^*)-{k}^\top \big(K+\sigma^2 I\big)^{-1}{k}.
 \end{align*}

\section{Learning-supported Model Predictive Force and Motion control}
\label{Chapter:learning_MPC}
We aim for precise and safe control of robot motions and interaction forces under constraint satisfaction. \rev{To do so, we use GPs for hybrid modeling of the system output \eqref{eq:nonlinear-system_output} while we assume negligible model plant mismatch in the system dynamics~\eqref{eq:nonlinear-system_stateequation} in combination with MPC. We exploit the fact that the "output" model uncertainty does not need to be propagated through the dynamics of the system and can be used for constraint tightening:}
In MPC, predictions are performed with the model. If the prediction model of the dynamical system is uncertain, i.e., contains a GP, uncertainty need to be propagated through the system \rev{and }the resulting distribution is typically non-Gaussian \cite{hewing2019cautious}. 
When the uncertainty only appears in the output projection, such uncertainty propagation is unnecessary. Therefore, the output variance, which reflects the uncertainty in the model, does not need to be approximated. The posterior of the GP forms a  Gaussian distribution over the full prediction horizon without necessarily growing over time or becoming increasingly uncertain. \rev{We show how this structure leads to a simplified stochastic MPC formulation and derive a tailored chance-constrained formulation.}

\rev{We start the discussion by considering that the model uncertainty is neglected in the MPC controller. Afterwards, we consider how the covariance of the GPs can be used for constraint tightening to achieve safe control via chance constrained MPC.}
 
 \subsection{Nominal model predictive control -- neglecting the uncertainty of the learned model}
  \label{sec:nomCase}
\rev{In practice, the model uncertainty is often neglected in the MPC predictions, i.e., a nominal MPC scheme with the learned model is used. The  (unknown) output $h$ in the prediction is replaced by the learned hybrid model $\tilde h( x)=h_\textup{fp}(x)+h_\textup{ml}(x)$. In case of the hybrid model outlined in Section \ref{sec:force_models} we use the mean value for the prediction. The optimal control problem \eqref{eq:OCP_pf} with the hybrid output model~\eqref{eq:hybrid_output} in constraint \eqref{eq:con_e_pf} remains otherwise unchanged.}

\rev{As a first step, we establish recursive feasibility and convergence with the learned output model, assuming that the GP matches reality ideally, i.e., $\mathbb{E}\left(\tilde h(x)\right)=h(x)$, and that no noise is present: $\sigma_y(x)=\sqrt{\kappa^+(x,x)}=0$. To do so, the GP part of the hybrid model must satisfy the conditions required for Theorem \ref{theo:pf}, i.e., Assumptions \ref{ass:compact_sets_xu}-\ref{ass:pf4}. Only Assumption~\ref{ass:Lipschitz_xy_unique}, specifically the smoothness of the output model, is critical. The required smoothness can be ensured, posing the following conditions on the GP part of the learned output map:}
\rev{
\begin{assumption}\label{ass:diff_gp}
The prior mean and covariance function $m:\Rn{x}\to \R$,  $\kappa:\Rn{x}\times\Rn{x}\to \R$, and the first-principles model part $h_\text{fp}:\Rn{\text{x}} \to \Rn{\text{y}}$ are continuously differentiable for all $x,x' \in \mathcal X$. 
\end{assumption}}
\rev{Provided that these conditions hold, we can establish the following lemma:
\begin{lemma}
Assumption~\ref{ass:diff_gp} implies that the learned output function $h: \Rn{\textup{x}} \to \Rn{\textup{y}}$ is continuously differentiable and locally Lipschitz for all $x\in \mathcal{X}$.
\label{theo:nominal_learned_MPC}
\end{lemma}}

\rev{\begin{proof}
The output function $h$ is given by $h(x)=\mathbb{E}(\tilde h(x))=h_\textup{fp}(x)+ m(x)
+{k}^\top (K+\sigma^2 I)^{-1}\big(\boldsymbol{\hat {F}}-\boldsymbol{m}(\boldsymbol{x})\big)$, where the entries of $K$ and ${k}$ are $K^{i,j}=\kappa(x^{i},x^{j})$ and ${k}^{i}=\kappa(x^i,x)$ with $i,j \in \{1,\ldots, n_{D}\}$ cf. \eqref{eq:hybrid_output} and \eqref{eq:mean}.
Differentiability of $h$ is guaranteed via differentiability of the summands. 
Local Lipschitz continuity of $h$ on each compact set $\tilde{\mathcal{X}}\subset \mathcal{X}$ follows since the restriction of any continuously differentiable function on a compact set is Lipschitz. 
\end{proof}}
\rev{\begin{corollary}
Provided that Assumptions \ref{ass:compact_sets_xu}~-~\ref{ass:diff_gp} hold.
Then the model predictive feedback resulting from repeatedly solving~ \eqref{eq:OCP_pf} using the nominal learned output model~\eqref{eq:hybrid_output} for prediction without model-reality mismatch and noise is recursively feasible, and the control error converges to zero.
\end{corollary}}
\rev{Common prior mean and covariance functions that allow satisfying Assumption \ref{ass:diff_gp} are the squared exponential covariance, periodic covariances, or specific parametrization of the Mat\'ern covariance \cite{rasmussen2006gaussian}.}

 \rev{\subsection{Ensuring safety -- output chance constraint model predictive control}}
 \label{sec:robustCase}
 \rev{Neglecting the uncertainty in the output does not allow to ensure constraint satisfaction. We outline how to use the  posterior variance to include a measure of the modeling errors into the controller and thus increase  safety. 
To do so, we reformulate the output constraints as chance constraints and tighten them by a set which includes most of the uncertainty. 
Doing so requires that the following assumption with respect to the error of the hybrid model holds:}
\begin{assumption}
There exists a finite bound ${b}_j: \tilde{\mathcal{X}} \to \R$ for each approximation error $\Big(h_j(x)-\mathbb{E}\big(\tilde h_j(x)\big) \Big)$ on a compact set $\tilde{\mathcal{X}}\subset \Rn{\textup{x}}$ such that
$
p\left(|h_j(x)-\mathbb{E}(\tilde h_j(x))|>{b}_j(x)\right)<1-\epsilon
$
for all $x \in \tilde{\mathcal{X}}$, $j\in \{1,\ldots,n_y\}$ and $\epsilon \in (0,1)$, where $p$ denotes the probability. 
\label{ass:true_approximationvar}
\end{assumption}
\rev{Deriving such bounds is, for example, outlined in \cite{koller2019learning} and in \cite{lederer2019uniform}.} In practice, often the posterior variance of a GP is assumed to approximate these error bounds ${b}_j$  directly, cf.  \cite{cao2017gaussian, soloperto2018learning, grancharova2008explicit}. 
It has been reported that this posterior variance can lead to an underestimation of the uncertainty, especially in multi-step-ahead predictions  \cite{hewing2020on}. 
In our setup, no 
uncertainty propagation is needed. 
Hence, we exploit multiples of the posterior standard deviation as a measure for the approximation quality. 
\rev{Including chance constraints for the output,} the optimal control problem~\eqref{eq:OCP_pf} is slightly modified. In particular the control error equation~\eqref{eq:con_e_pf} is changed into 
\begin{align}
\label{eq:mean_errors}
   e_\text{pf}(\tau) &= r_\text{pf}\left(l( z(\tau))\right)-\mathbb{E}\left(\tilde h\left( x(\tau)\right)\right) .  
\end{align}

\begin{figure}[t]
   \centering      
      \subfloat[Based on the posterior variance (gray area) of the output model the original constraints (red dashed line) are tightened.   ]
      {\input{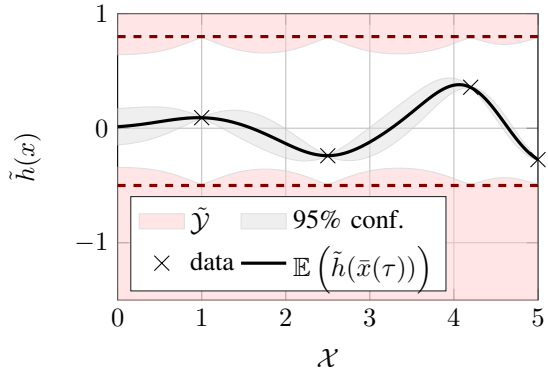}}\quad
      \subfloat[Based on the maximum value $\sigma_{y,\text{max}}$ of the posterior variance (gray area) the original constraints (red dashed) are tightened. ]
      {\input{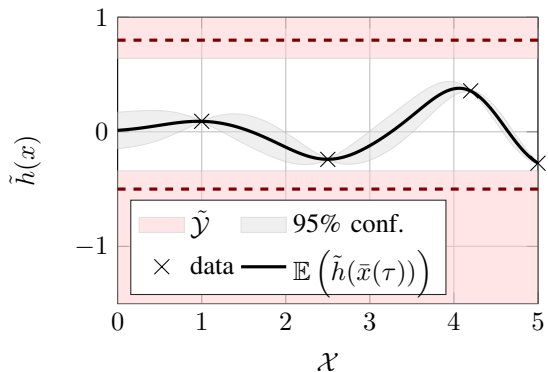}}    \quad 
   \caption[]{Constraint tightening.}
   \label{fig:constraint_tigthening}
\end{figure}

 This way, the control error and objective function are deterministic. Also, the output constraints in \eqref{eq:con_X_pf} are altered into chance constraints
$p\left(\tilde h( x(\tau)) \in  \mathcal{Y}\right)\geq p_\mathcal{Y},$
where $p_\mathcal{Y}\in (0,1)$ denotes a chosen probability for the satisfaction of the constraints. 
This chance constraint can be reformulated into 
\begin{equation}
\mathbb{E}\left(\tilde h( x(\tau))\right) \in \tilde{ \mathcal{Y}},
\end{equation}
where the modified constraint set $\tilde{ \mathcal{Y}}$ is defined via
$
\tilde{ \mathcal{Y}}:=\mathcal{Y}\ominus \mathcal{R}.
$
The set $\mathcal{R}$ is constructed  via the respective confidence level or error bound that belong to the desired reliability $p_\mathcal{Y}$.  
For example, $\mathcal{R}=[-2\sigma_y(x), 2 \sigma_y(x)]$ for a one dimensional output with $p_\mathcal{Y}\approx 95.45$, where $\sigma_y$ is the standard deviation of the learning based output $\tilde h(x)$. 
Alternatively, the constraint tightening can consider the worst case realization of the uncertainty over a compact set $\tilde{\mathcal{X}}$ such that $\sigma_{y,\text{max}}:=\max\limits_{x\in \tilde{\mathcal{X}}} \sigma_y(x)$ is used in the construction of $\mathcal{R}$ instead of $\sigma_y$. 
Figure~\ref{fig:constraint_tigthening} illustrates the constraint tightening. 
In both cases, we need to ensure \rev{the existence of solutions and (initial)} feasibility of the  optimal control problem with tightened constraints. Therefore, we rely on the following assumptions.
\rev{\begin{assumption}
The tightened output set $\tilde{\mathcal{Y}}$ is closed. 
\label{ass:shrink_Y}
Furthermore, the intersection of the state constraint set $\mathcal X$ and the preimage $ h^{-1}\left(\tilde{\mathcal{Y}}\right)$ of the tightened output constraint set $\tilde{\mathcal{Y}}$ is nonempty and contains the reference $r_\textup{pf}$.\label{ass:pre_shrink_Y}
\end{assumption}}


When considering the uncertainty in the output via tightened constraints, the following \rev{Output Chance Constrained MPC setup is used:\\
\it{Output Chance Constraint MPC:}
  In the repeatedly solved optimal control problem ~\eqref{eq:OCP_pf}, the learned output model  $\tilde h( x)=h_\textup{fp}(x)+h_\textup{ml}(x)$ is used in the control error formulation $e_\textup{pf}$ ~\eqref{eq:mean_errors}. Furthermore, the tightened output constraint set $\tilde{\mathcal{Y}}$ instead of $\mathcal{Y}$ is used in   \eqref{eq:con_X_pf} and  \eqref{eq:term_con_pf}.}

\rev{In comparison to the nominal case, alterations in the output constraints are performed. 
In contrast to classical stochastic MPC, the state constraints remain untouched by the uncertainty. Hence, stability/convergence results can be established similarly as in Section~\ref{sec:nomCase}. The assumption on the closedness of $\tilde{\mathcal{Y}}$ is the only assumption at stake, which is, however, guaranteed by  Assumption~\ref{ass:shrink_Y}. Moreover we pose additional conditions on the nonemptyness of the intersection of $\tilde{\mathcal{Y}}$ and $ h^{-1}\left(\tilde{\mathcal{Y}}\right)$ which is necessary for an (initial) feasible solution to exist.
}
\begin{observation}
Given that Assumptions~\ref{ass:compact_sets_xu}- \ref{ass:pre_shrink_Y} hold,
 \rev{then the output chance constrained} predictive controller is recursively feasible with the probabilistic satisfaction of the constraints and the control error $e_\textup{pf}$ \rev{from \eqref{eq:mean_errors}} converges to zero. 
\end{observation}

\rev{Note that the} constraint satisfaction of the true outputs $h(x)$ with respect to $\mathcal Y$ can be guaranteed  
\rev{only} with  probability  $p_\mathcal{Y}$. 

\section{Application example}\label{sec:example}
\begin{figure}[t]
\centering
\input{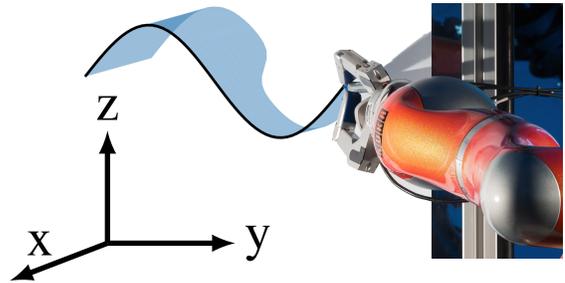}
\caption{Illustration of the base frame and the reference for the robot writing task. }
\label{fig:extended_path}
\end{figure}

The benefits of a learning-supported MPC force controller are shown here on a robotic application example.  
The robot should follow a Cartesian reference path. 
At the same time, the robot is in contact with a flexible surface and should apply a desired force along the Cartesian reference normal to the contact surface. 
As an illustrative example, we use a calligraphic writing task, where different contact forces along the reference are desired. 
The y- and z-axes of the base frame  are parallel to the whiteboard surface, while the x-axis is normal to the surface, see also Fig.~\ref{fig:extended_path}. The position reference is a sinusoidal curve depicted as a black line in Fig.~\ref{fig:extended_path} and the force reference is indicated via a blue shaded area along the x-axis. 
Since the reference consists of three dimensions, we consider the control of three joints of the robot to obtain a square input-output structure $n_\text{y}=n_\text{u}=3$.

\subsection{Robot and force model}
\label{sec:force_models}
The derivation of the dynamical system equation based on first principles is given in Appendix~\ref{app:system_model}. In this subsection, the hybrid output modeling is discussed in more detail. 
The system outputs are given via 
 $y=(p_\text{e,y},p_\text{e,z},F_\text{n}, z_1)^\top$, where $p_\text{e,y}$ and $p_\text{e,y}$ are the Cartesian end-effector (pen tip) position in y- and z-direction. 
 These are obtained via the direct kinematics of the robot. 
 The third output $F_\textup{n}\in \R$ denotes the normal contact force between the pen and the whiteboard in x-direction of the base frame and $\theta=z_1$  represents the path parameter. 
 We assume that sufficiently precise first-principle output models for $p_\text{e,y},p_\text{e,z}$ exist. 
 In contrast, the contact forces also depend on the robots surroundings and can vary in their complexity arbitrarily.  
 Therefore, the modeling of the contact forces is performed with a hybrid model. 
 We use a simple first-principle models to capture the basic physical knowledge and enhance extrapolation. It is supported by a Gaussian process to increase the model accuracy and flexibility. 
 To underline the benefits of the hybrid model (increased model accuracy and uncertainty quantification for robustification), three different modeling schemes are investigated and compared in the following.

  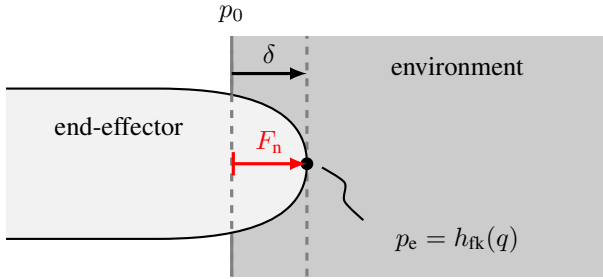
\begin{figure}[t]
\centering
\begin{tikzpicture}[thick]
\coordinate (surf1a) at (3,-1.5); 
\coordinate (surf1b) at (3,1.7); 
\coordinate (surf2a) at (4,-1.5); 
\coordinate (surf2b) at (4,1.7);
\coordinate (surf3a) at (8,-1.5); 
\coordinate (surf3b) at (8,1.7);

\draw [draw=none, fill=gray!40!white,very thick] (surf1a) -- (surf1b) -- (surf3b) -- (surf3a)--cycle;

\draw [gray,solid, very thick] (surf1a) -- (surf1b);
\draw [gray,dashed,  very thick] (surf2a) -- (surf2b);

\draw [black,solid,-latex, very thick] ($(surf1b)-(0,0.5)$) --node[above]{$\delta $} ($(surf2b)-(0,0.5)$);

\draw[black,fill=gray!10!white] (0,1) -- 		(2,1) 
 to[out=0,in=90] (4,0) to[out=270,in=0] (2,-1) -- 				(0,-1);

\draw [gray,dashed, very thick] (3,-1.5 ) -- (3,1.5 );
\filldraw (4,0) circle (2pt);

\draw [black] plot [smooth] coordinates {(4.2,-0.05) (4.5,-0.25) (4.5,-0.5) (4.75,-0.75) };

\draw [red,solid,|-latex, very thick] (3,0) --node[above]{$F_\text{n}$} (4,0);

\node []at (1.5,0.5) {end-effector};
\node []at (6,1.3) {environment};
\node []at (6,-1) {$p_\text{e}=h_\text{fk}(q)$};
\node []at (3,2) {$p_0$};

\end{tikzpicture}
\caption{The penetration depth $\delta$ is the distance between the end-effector pose $p_\text{e}=h_\text{fk}(q)$ and the initial contact position $p_0$ along the normal of the environment surface. The resulting normal force $F_\text{n}$ is modeled as a function of $\delta$.  }
\label{fig:penetration_depth}
\end{figure}

Among the purely elastic models, which can be used to quantify the interaction, are the linear spring model of Hook 
\begin{equation}
F_\text{n,Hk}:=h_{F_\text{n,Hk}}(x)=K_\text{e} \delta 
\label{eq:force_Hk}
\end{equation} 
and the nonlinear spring model of Hertz
\begin{equation}
F_\text{n,Hz}:=h_{F_\text{n,Hz}}(x)=K_\text{e} \delta ^\alpha
\label{eq:force_Hz}
\end{equation} 
where $K_\text{e}\in \R$ is a spring constant and the coefficient $\alpha \in \R_0^+$ introduces nonlinearity and is equal to 1.5 in the original work of Hertz. The penetration depth $\delta$ is shown in Fig.~\ref{fig:penetration_depth}.
As the third model, we propose a hybrid model composed of a linear spring model and a GP via $h_{\text{ml}}(x)\in \R$. It is given by 
\begin{equation}
F_\text{n,HkGP}:=h_{F_\text{n,HkGP}}(x)= K_\text{e} h_\delta (q)+h_{\text{ml}}(x).
\label{eq:force_HkGP}
\end{equation} 

\begin{figure*}[ht]
\centering
\includegraphics[width=\textwidth]{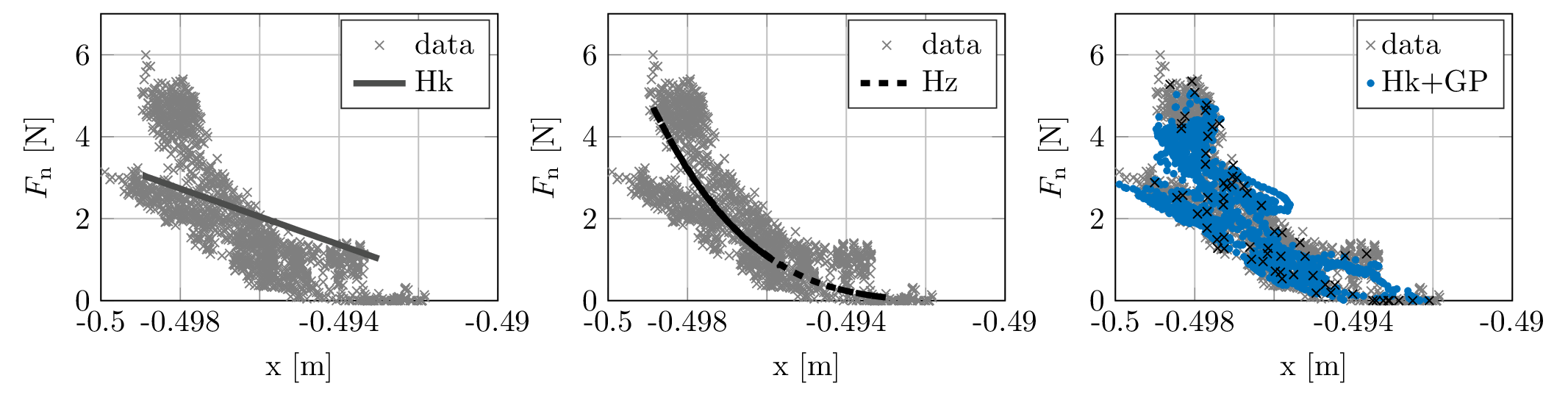}
\includegraphics[width=\textwidth]{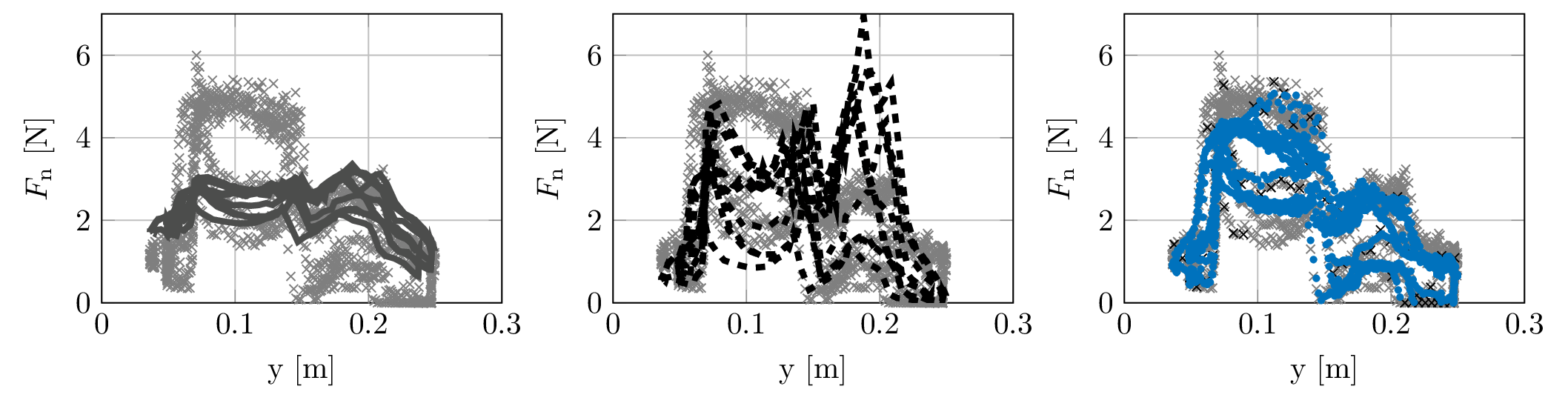}
\includegraphics[width=\textwidth]{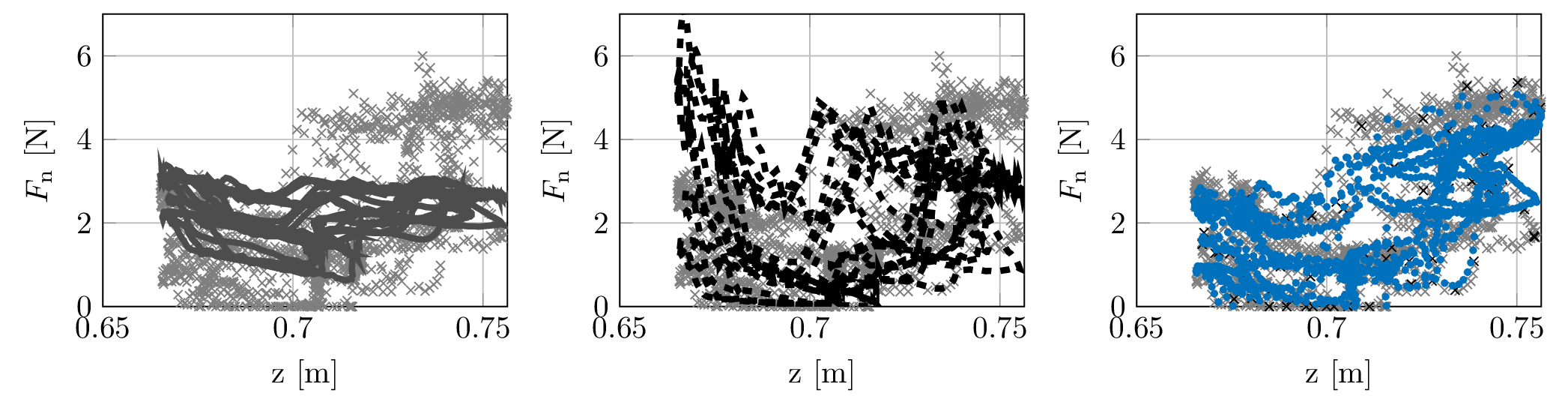}
\caption{Comparison of linear (left), nonlinear (middle), and hybrid force model (right) plotted over the Cartesian x-direction (top), y-direction (middle), and z-direction (bottom).}
\label{fig:force}
\end{figure*}

\subsubsection*{\rev{Training data}}
The parameters of the first-principle models are estimated on data that was collected by a baseline controller. This baseline controller is outlined in \cite{matschek2017force} for a similar control task.   
Multiple runs of the controller are performed in the vicinity of the original reference.  
The full available data set consists of 20000 data points, where each data point represents the measured joint angle positions and the corresponding contact force, as depicted in Fig.~\ref{fig:force} as gray crosses.
 The training data used for the GP consisted of 85 data points, which is a subset of the evaluation data set. These points are shown in Fig.~\ref{fig:force} (right hand side) as black crosses. 
 They have been chosen with \SI{0.015}{rad} minimum Euclidean distance in terms of the angular positions between each other to obtain an equally spread coverage of the considered space. 
 
 \subsubsection*{\rev{Modeling  and learning results}}
 Fig.~\ref{fig:force} plots from top to bottom the measured and modeled forces over the Cartesian x-, y-, and z-directions of the base frame. 
 Additionally to the data, the results for the linear spring (Fig.~\ref{fig:force}, left), the nonlinear spring (Fig.~\ref{fig:force}, middle) and the hybrid model (Fig.~\ref{fig:force}, right) are depicted. 
 An optimization-based identification of the parameters resulted in $K_e=341.56$ for the linear, and $K_e= 2.5276\cdot 10^{8}$, $\alpha=3.7651$ for the nonlinear model. 
 The latter was parameterized using a two-stage identification procedure\cite{diolaiti2005contact}. 
 The hybrid model combines the linear model with a Gaussian process with zero prior mean and squared exponential prior covariance function. 
 Since, the board surface is assumed to be placed parallel to the y-z-plane of the robots base frame, the penetration depth is defined along the x-axis with a assumed initial contact position of {-0.49}{m}.

 As can be seen in Fig.~\ref{fig:force} (left), the linear model (solid gray line) is a poor approximation of the measured forces. 
 The root mean square error between the linear force model output and the data is  \SI{1.06}{N}. 
 The  nonlinear force model (black dashed line) approximates the interaction forces better than the linear one, cf. Fig.~\ref{fig:force}, middle. The corresponding root mean square error is \SI{0.91}{N}. 
 The hybrid model can also take the non-perfect alignment of the whiteboard as well as small unevenness and stiffness changes of it into account. This can also be seen in Fig.~\ref{fig:force}, middle and bottom. While the contact forces are often underestimated by the linear model, the nonlinear spring model sometimes overestimates them, see Fig.~\ref{fig:force} (center) in $y\in\left[0.16,0.22\right]$ or Fig.~\ref{fig:force} (bottom middle) in $z\in\left[0.66,0.7 \right]$. In contrast, the hybrid model nicely fits the data without overfitting its noise. It shows a root mean square error of \SI{0.41}{N}, which corresponds to an error reduction of  \SI{61}{\%} compared to the linear spring model and of \SI{55}{\%} compared to the nonlinear model.

\subsection{Controller setup}
\label{sec:control_params}

\begin{figure}[t]
\centering
\input{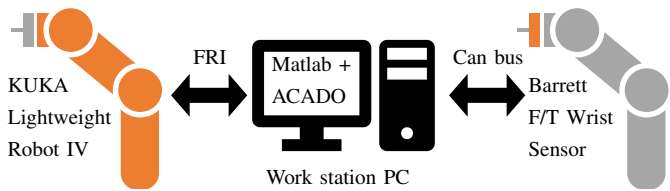}
\caption{Linux Work station PC with Ubuntu 12.04 and an Intel Xeon(R) X5675 processor with 3.07 GHz x6 connected to the KUKA robot via the fast research interface (FRI) and to the force/torque sensor via can bus.
}
\label{fig:setup}
\end{figure}

Using the hybrid model, a model predictive controller is implemented. The simulations of the system are performed in MATLAB and the optimal control problem is solved using ACADO \cite{acado}.
In the experiments, the fast research interface establishes the communication between the KUKA lightweight robot and the work station PC  \cite{schreiber2010fast}. 
\rev{Joint torques are sent over the interface as inputs to the robot and the robot's inbuild joint position and velocity sensors are used to measure the states which are sent back to the PC. } 
A CAN-bus builds the connection to the wrist force-torque sensor \rev{\cite{barrett2020}, which provides the contact force data}. The communication interface for the control of the robot from MATLAB was designed in \cite{bargsten2013modeling}. The sensor interface is outlined in \cite{bethge2016master}. 
\rev{Figure~\ref{fig:setup} describes this overall setup.}
For the following simulations and experiments the same cost function and control parameters are used, which are given in Appendix~\ref{app:params}.

\begin{figure}[t]
\centering
\includegraphics[width=\columnwidth]{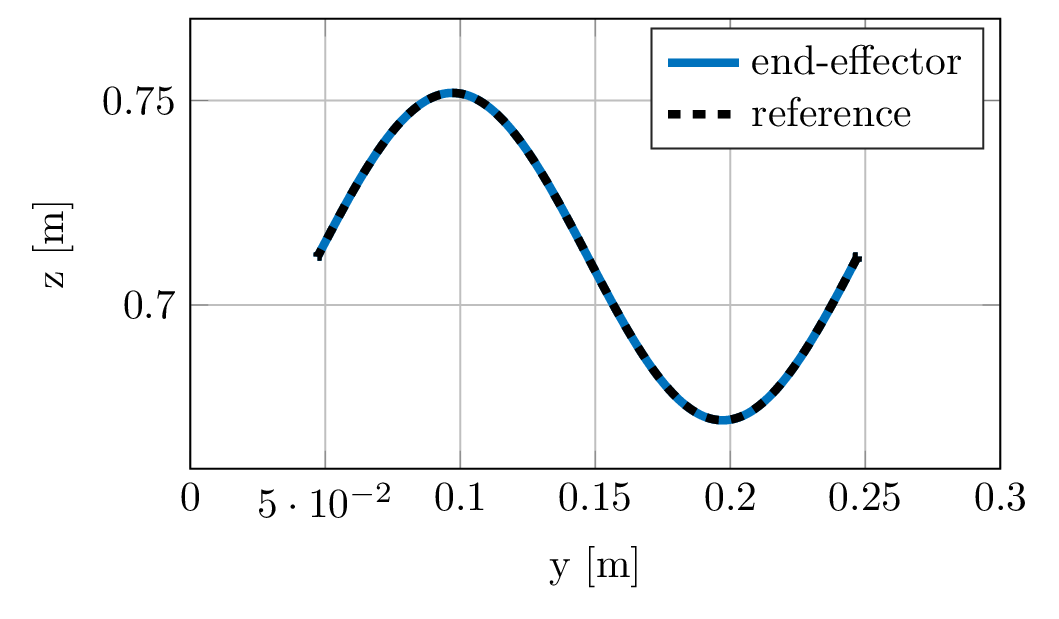}
\caption{Cartesian end-effector position in simulations.}
\label{fig:sim_HkGP_out_yz}
\end{figure}

\begin{figure}[t]
\centering
\includegraphics[width=\columnwidth]{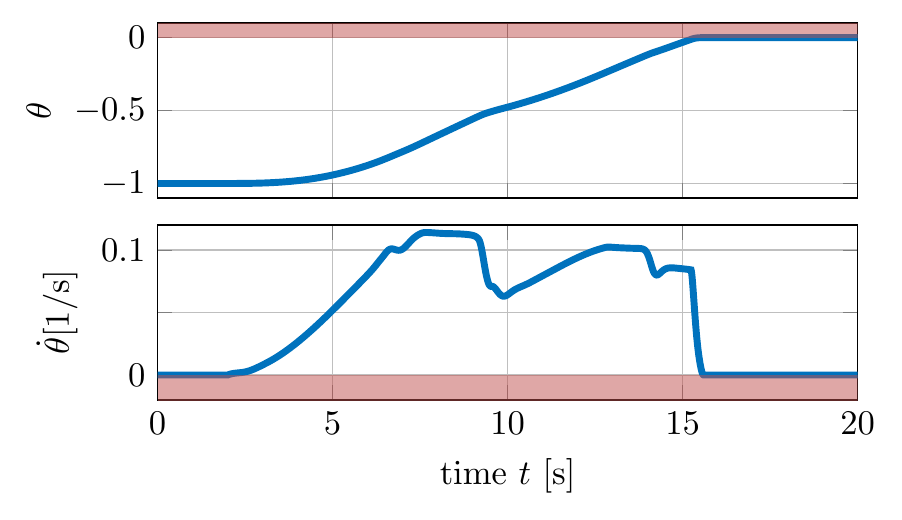}
\caption{Virtual system in simulation. \rev{The path parameter $\theta$ (top) is constrained to non-positive values (red area depicts constraints) while the  path speed (represented by $\dot \theta$, bottom) is restricted to non-negative values for forward motion.} }
\label{fig:sim_HkGP_out_dist_theta}
\end{figure}

\begin{figure}[t]
\centering
\includegraphics[width=\columnwidth]{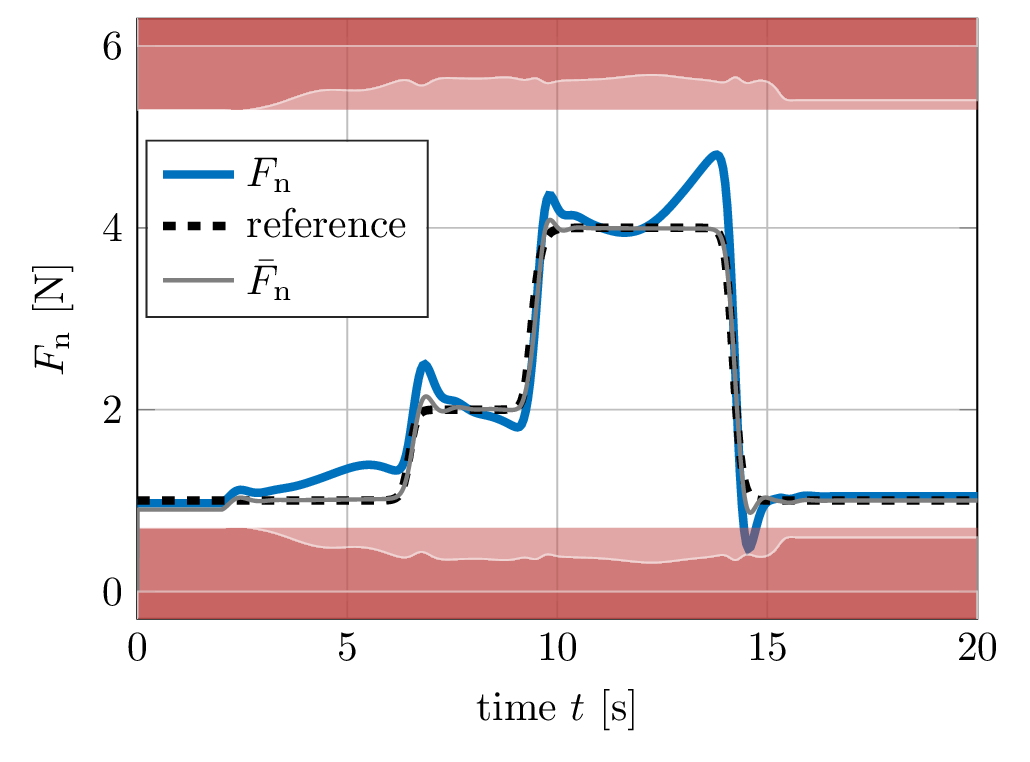}
\caption{Contact force in simulation. \rev{Simulated force (solid blue line) and predictions inside the optimal control problem (thin gray solid line). The dark red area depicts original constraints, medium red shows constraint tightening based on state dependent variances, light red area depicts shrunken constraints for state-independent over-approximation of uncertainty. }}
\label{fig:sim_HkGP_out_F}
\end{figure}

\begin{figure*}[t]
\centering
\includegraphics[]{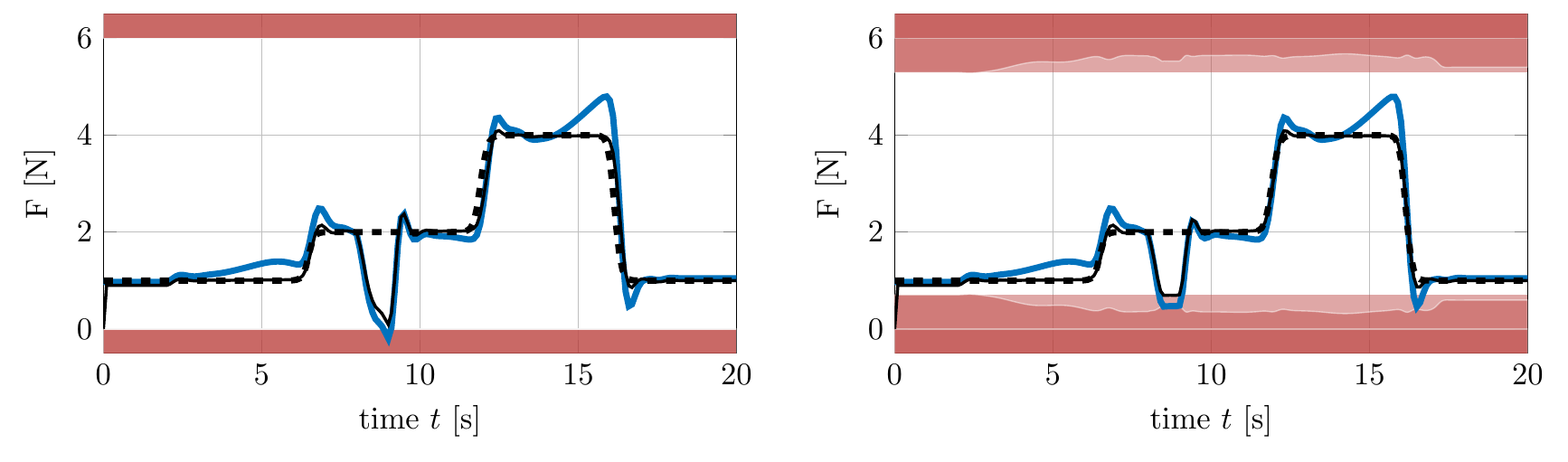}
\caption{Comparison of control performance for using original constraints and tightened constraints when disturbance occurs. }
\label{fig:sim_HkGP_out_dist_Fyz}
\end{figure*}

\subsection{Simulation results}

\subsubsection*{Learning-supported hybrid force control simulation}

First simulations are performed following a reference path.
In these simulations, no model-plant mismatch in the dynamic equations is assumed. 
Instead, a simulated mismatch in the contact forces is considered. 
To do so, a purely data based force model using 515 data points was trained. Since this model is using more than 6 times as much data as the hybrid GP, it gives an even better representation of the true contacts. 
Note that the computational demand for GP inference grows in general cubically with the amount of data points. Thus, such a big GP model is neither suited for fast online simulations nor for the use in a real-time optimal controller. Nevertheless, it allows us to test model-plant mismatch in simulations.

The Cartesian end-effector position in this simulation in the y-z-plane can be seen in Fig.~\ref{fig:sim_HkGP_out_yz}. The reference path is shown in black dashed line, while the simulated robots position is depicted as solid blue line.
To achieve an optimal tracking performance, the model predictive controller is adjusting the reference evolution. The virtual system  states are depicted in Fig.~\ref{fig:sim_HkGP_out_dist_theta}. The path parameter $\theta$ starts at $-1$ and moves toward its end value $0$ without exceeding it. The non-admissible area, i.e., the area where the constraints are violated, is depicted as a red-shaded area. The path parameter derivative is shown in Fig.~\ref{fig:sim_HkGP_out_dist_theta}, bottom. 
As can be seen, the path-following control formulation leads to a nontrivial velocity profile. It \rev{enables} optimal path following while considering limitations, e.g., in the joint velocities.

The evolution of the contact force over time is depicted in Fig.~\ref{fig:sim_HkGP_out_F}. 
The model-plant mismatch in the output equation leads to a deviation of the achieved contact force from its reference, cf. Fig.~\ref{fig:sim_HkGP_out_F}. 
Even though the MPC prediction model claims that the contact force is close to the reference (thin gray line), the true contact force (blue solid line) deviates from it. 
The original force constraint (dark red area) might be violated due to this model-plant mismatch. To cope for this model-plant mismatch the original force constraint is adjusted using the posterior variance of the GP.
 The reliability of the hybrid GP model along the motion is calculated via the 2$\sigma_y$ confidence bound. It is tightening the original constraint  [\SI{0}{N} \SI{6}{N}] , which is indicated by the medium red area. An over-approximation of this uncertainty related tightening is performed, which is indicated by the light red area in Fig.~\ref{fig:sim_HkGP_out_F}. The resulting robustly tightened box constraints are [\SI{0.7}{N} \SI{5.3}{N}]. 
The MPC controls the system such that the force model predictions stay inside the feasible area (white space). Hence, the true contact force (blue solid line) stays inside the original constraints of [\SI{0}{N}~\SI{6}{N}] with a chosen probability. 


\begin{figure*}[t]
\centering
\includegraphics[]{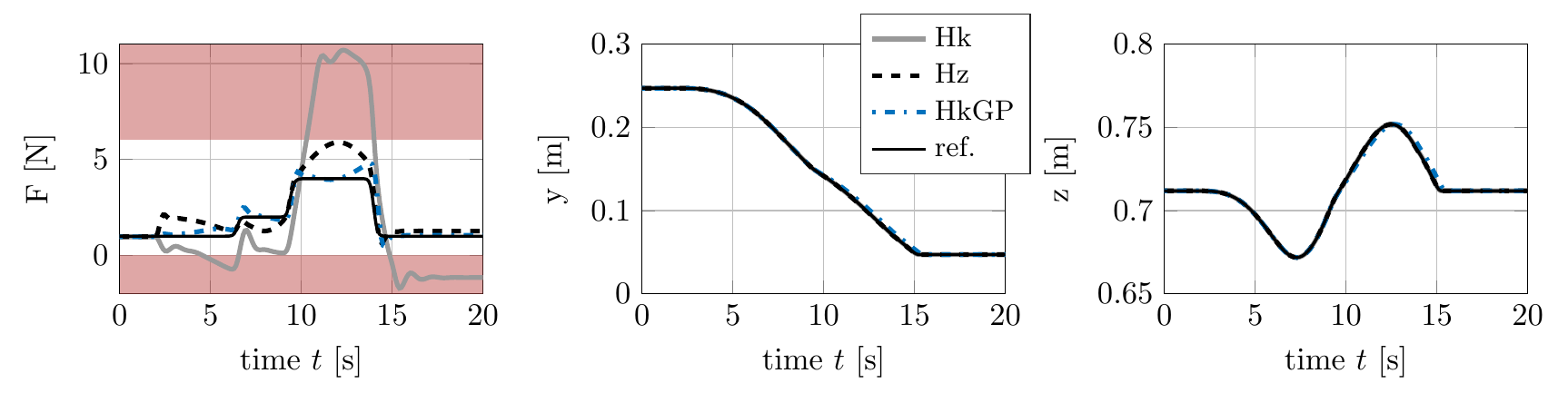}
\caption{Comparison of controllers with first-principle and hybrid force model in simulation.}
\label{fig:sim_comp_out_Fyz}
\end{figure*}

\begin{figure*}[t]
\centering
\includegraphics[]{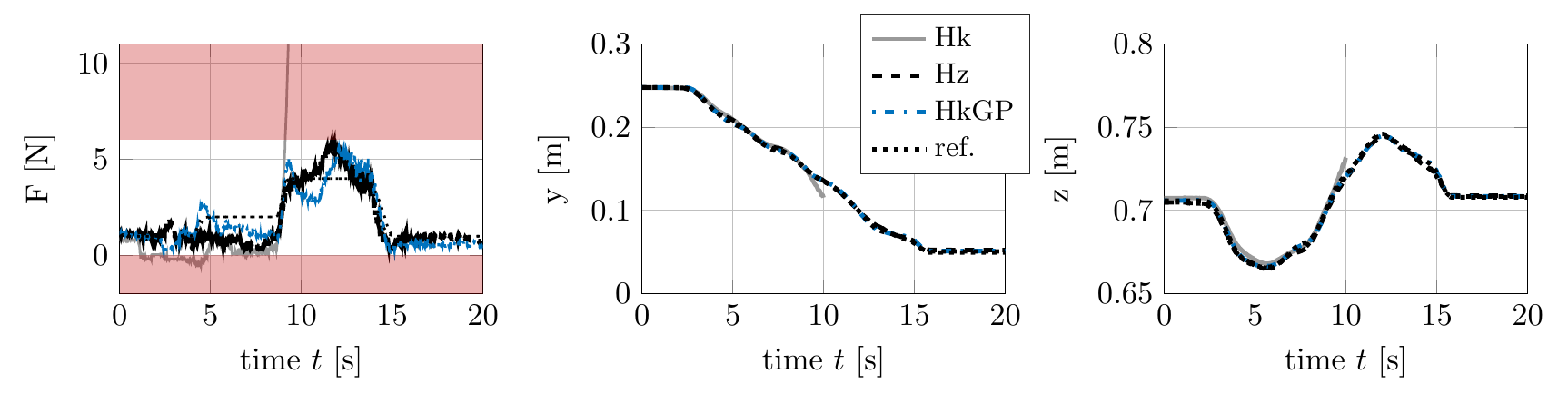}
\caption{Comparison of controllers with first-principle and hybrid force model in experiment.}
\label{fig:exp_comp_out_Fyz}
\end{figure*}

\subsubsection*{Disturbance rejection} 
Additionally, a disturbed case is considered to underline the effect of the constraint tightening. 
In this case an additive input disturbance at the fourth joint of \SI{-1.3}{Nm} occurs between $t=\SI{8}{s}$ and $t=\SI{9}{s}$. 
The effect on the contact force is illustrated in Fig.~\ref{fig:sim_HkGP_out_dist_Fyz}. 
The resulting contact force without constraint tightening is depicted in the left subplot. 
Additionally, the disturbed simulation results with constraint tightening are plotted  in Fig.~\ref{fig:sim_HkGP_out_dist_Fyz} right. 
For both cases, the prediction used inside the controller is plotted as black thin line. 
As can be seen in Fig.~\ref{fig:sim_HkGP_out_dist_Fyz}, the disturbance pushes the robot away from the board between $t=\SI{8}{s}$ and $t=\SI{9}{s}$, which leads to smaller forces.
The hybrid GP model (thin black line) overestimates the contact forces in this area, such that it calculates minimum forces of \SI{0.1}{N} in the untightened case. However, the simulated force of the large GP (blue line) is falling below zero, i.e., indicating that the robot has lost contact. Hence, the original lower constraint of \SI{0}{N} is violated, due to the model-plant mismatch.
In contrast, the tightened constraints prevent from this contact loss, cf. Fig~\ref{fig:sim_HkGP_out_dist_Fyz} right. They become active at around $t=\SI{8.5}{s}$, limit the occurring force error, and introduce a safety margin such that the true force (blue line) obeys the original force constraints, cf. Fig.~\ref{fig:sim_HkGP_out_dist_Fyz}, right. 
Hence, a safe and stable contact between the robot and the whiteboard is guaranteed at least with the chosen probability despite the occurring disturbance.

\subsubsection*{Comparison to first-principle model-based  controllers}

Additional simulation studies are performed to compare the closed loop control performance of the learning-supported controller with controllers using the first-principle models. 
To do so, the same controller parameters are used as listed in Section~\ref{sec:control_params} for all simulations. 
The difference between the three controller setups compared in this section lies mainly in the force output model. 
The linear first-principle model, the nonlinear first-principle model and the hybrid model from Section~\ref{sec:force_models} are used, respectively. 
Fig.~\ref{fig:sim_comp_out_Fyz} shows the system outputs for the linear spring model (Hk), the nonlinear force model (Hz) and the hybrid model (Hk+GP). 
As can be seen, the influence of the different force models on the performance in the position controlled subspace (Cartesian x- and y-direction) is comparably small. 
A clear difference between the controller performances can be seen in the force controlled output, cf. Fig.~\ref{fig:sim_comp_out_Fyz} left. 
The model-plant mismatch in the linear case (depicted as gray solid line) is so large, that a constraint violation of the maximum and minimum force limit occurs.  
In the nonlinear case, all constraints on the contact forces are satisfied and the root mean square force control error is \SI{0.77}{N}. 
Still, the hybrid case outperforms the nonlinear force model MPC, cf. Fig.~\ref{fig:sim_comp_out_Fyz}.  The controller with hybrid model shows a root mean square error for the force of \SI{0.25}{N}. 
This corresponds to a force error reduction of \SI{91.3}{\%} compared to the MPC with linear force model and \SI{67.5}{\%} with respect to the MPC with nonlinear first-principle force model.

\subsection{Experimental validation}

\begin{figure}[t]
\centering
\includegraphics[width=\columnwidth]{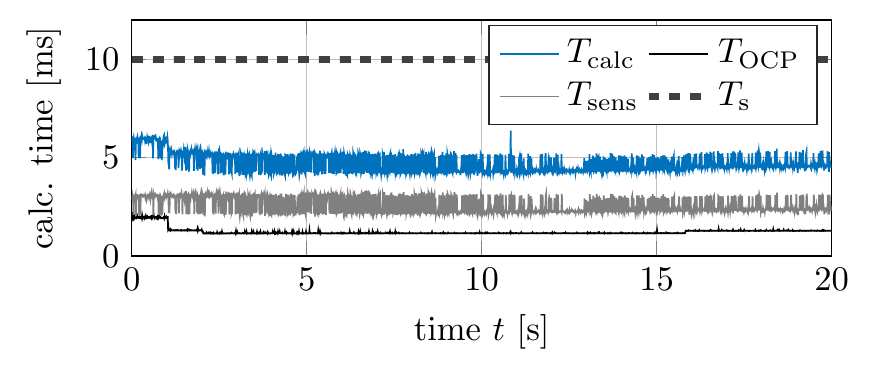}
\caption{\rev{Total calculation time $T_\text{calc}$ (blue) in each sampling instance and its main components, the sensor communication delay $T_\text{sens}$ and the calculation time of the optimal control problem $T_\text{OCP}$.}}
\label{fig:calc}
\end{figure}

\rev{Figure~\ref{fig:calc} underlines that the controller is real-time feasible, showing the computation time of the experimental implementation. Real time feasibility is given due to the total calculation time $T_\text{calc} $ (blue) being smaller than the sampling time of $T_\text{s}=\SI{0.01}{s}$} at all times.
The controlled system outputs for the experimental validation of the learning-supported model predictive controller are shown in Fig.~\ref{fig:exp_comp_out_Fyz} in blue dash-dotted lines.
 For comparison, the MPC with linear and nonlinear spring force models are depicted in  Fig.~\ref{fig:exp_comp_out_Fyz} in gray solid and black dashed lines, respectively. 
Compared to the simulation results, an additional model-plant mismatch in the dynamics as well as additional noise in the sensor readings occur. Therefore, all controllers show larger control errors in the experiments compared to the simulation studies.
Due to the large model-plant mismatch in the linear force model, the controller even became unstable, cf. Fig.~\ref{fig:exp_comp_out_Fyz} gray line. 
The MPCs with nonlinear and hybrid force model perform better than the linear case. 
Their performance in the position controlled subspace is comparable, see Fig.~\ref{fig:exp_comp_out_Fyz} (middle and right). 
Comparing their force control errors shows that the learning-supported controller shows around \SI{12}{\%} smaller maximum errors and around \SI{14}{\%} smaller average control errors. 
Hence, the superior force tracking performance of the learning-supported controller over the first-principle based MPC is not only valid in simulations but also in experiments. 
Moreover, the MPC with nonlinear first-principle force model violates the maximum force constraints, see black line in Fig.~\ref{fig:exp_comp_out_Fyz} (left). The GP not only approximates the real forces more accurately, but also \rev{enables} the variance-based constraint tightening for increased safety.  

\section{Conclusion}
This paper proposes a learning-supported model predictive controller for hybrid position and force control.
\rev{It exploits that the stochastic uncertainty affect only the static model parts, e.g., mappings from states to the controlled variables, and not the dynamics itself.}
Driven by the increasing demand for sensitive, and interactive, yet safe robots, we propose a model- and data-based controller that directly controls the forces of the robot along with its motion. 
The model predictive controller considers directly  constraints on inputs, states, and outputs. This way, constraints for positions, motions, and forces are satisfied. A path following predictive controller is used, where the reference speed is adjusted during runtime to provide an additional degree of freedom to the MPC. 
A Gaussian process supports the controller with learned hybrid force models.
The presented approach achieves improved approximation quality compared to standard first-principle force models and additionally improves the closed-loop system performance.
Moreover, additional robustness is achieved by using the \rev{stochastic} uncertainty description of the Gaussian process force model. 
It builds the basis for constraint tightening that allows for safe and reliable contacts even in case of disturbances and model-plant mismatch. 

Further research focuses on improving the controller performance via online updates of the force models. 
Furthermore, changing environmental conditions, such as moving objects or cooperating robots, should be addressed in future research. For example, learning from past measurements/batches might allow to build up an increasing  set of possible models that allow  transfer\rev{ing} and adapt\rev{ing} knowledge between different  situations. 
\rev{Moreover, online adaptation and learning is promising, yet leads to challenges such as a priori unknown constraints that we aim to address in future works, for example using the approach presented in \cite{batkovic2020safe}.}

\appendices

\section{Model of the considered light-weight robot}
\label{app:system_model}
The model of the robotic manipulator is derived based on first principles using the Lagrangian formulation \cite{siciliano2010robotics} leading to
\begin{eqnarray*}
\left(\begin{smallmatrix}
\dot x_1\\ \small{\vdots} \\\dot x_{n_\text{q}} \\  \hspace{-3pt} \dot x_{n_\text{q}+1}  \hspace{-3pt}\\ \vdots \\ \dot x_{2n_\text{q} \hspace{-3pt}}
\end{smallmatrix} \right)\hspace{-3pt}= \hspace{-3pt} 
\underbrace{
\left(\begin{smallmatrix}
 x_{n_\text{q}+1}						\\ 
 \vdots 						\\ 
 x_{2n_\text{q}}						\\  
  \\
 \hspace{-3pt}B^{-1} (x_1, \mydots, x_{n_\text{q}})  
 \left( 
 u\hspace{-2pt}-\hspace{-2pt} J^\top(x_1, \mydots, x_{n_\text{q}} ) F  \hspace{-2pt}-\hspace{-2pt}N(x) \right) \hspace{-3pt} \\
  \phantom{B}  
\end{smallmatrix} \right)
}_{f(x,u)}\\
 N(x)\hspace{-2pt}=\hspace{-2pt}C(x)\hspace{-1pt} \left(\begin{smallmatrix}\hspace{-1pt} x_{n_\text{q}+1} \hspace{-1pt}\\ \vdots \\ x_{2n_\text{q}} \end{smallmatrix} \right)
\hspace{-2pt}+\hspace{-1pt}\tau_\text{f}(x_{n_\text{q}+1}, \mydots, x_{2n_\text{q}})
 \hspace{-2pt}+\hspace{-1pt}\tau_\text{g}(x_1, \mydots, x_{n_\text{q}}).
\end{eqnarray*}
Here, $x_1,\ldots, x_{n_\text{q}}$ denote the joint angles, while $x_{n_\text{q}+1},\ldots, x_{2n_\text{q}}$ represent the joint angle velocities. For the example, we use joint  one, two and four of the lightweight robot depicted in Figure~\ref{fig:robo_writing}. Hence, $n_\text{q}=3$ and $x=(q_1,q_2,q_4,\dot q_1, \dot q_2, \dot q_4)^\top$, where $q_i$ denotes the angle of the $i$\textsuperscript{th} joint.
The configuration dependent inertia matrix of the robot is denoted by $B: \Rn{\text{q}} \to \mathbb{R}^{n_\text{q} \times n_\text{q}}$.
The input $u$ represents the joint actuation torques. 
The Coriolis and centrifugal effects are captured by $C: \Rn{\text{q}}\times \Rn{\text{q}} \to \mathbb{R}^{n_\text{q} \times n_\text{q}}$. 
While the gravitational force $\tau_\text{g}: \Rn{\text{q}} \to \Rn{\text{q}}$ for a rigid link manipulator depends only on joint positions, the friction torque $\tau_\text{f}: \Rn{\text{q}} \to \Rn{\text{q}}$ can be modeled via viscous and Coulomb friction such that is depends on joint velocities.  
We consider contact of the robot with the environment at the end-effector. The contact forces and moments occurring at the end-effector in a three-dimensional Cartesian space are captured by $F \in \mathbb{R}^6$. They can be mapped to the corresponding joint torques via $\tau_\text{ext}=J(q)^\top F$ with the manipulator Jacobian $J: \Rn{\text{q}} \to \mathbb{R}^{6 \times n_\text{q}}$. 
The model parameters are taken from \cite{bargsten2013modeling}, supplemented by the modifications presented in \cite{bethge2016master} to account for the used force-torque sensor.  
\rev{The standard deviation $\sigma$ of the noise  is approximately \SI{0.0388}{N}.}
This dynamical model is used for simulations as well as for prediction in the MPC. In contrast to \cite{bargsten2013modeling}, $N(x)=0$ is considered, as we use gravity compensation provided by the internal KUKA controller. Furthermore, friction and Coriolis effects only have a minor influence for small velocities, as considered here. Finally, a compensation of external torques, e.g., due to the environment contact, outside of the predictive controller is added to the optimal inputs. For the virtual system dynamics, a double integrator $\dot z_1=z_2$, $\dot z_2=v$ is used with virtual input $v$ and virtual output $z_1=\theta$.

\section{Controller parameters} 
\label{app:params}

A quadratic cost function is used, where:  
$ L_\textup{pf}=(e_\textup{pf}^\top,\theta) Q (e_\textup{pf}^\top,\theta^\top)^\top + (u^\top,v) R (u^\top,v)^\top$, $E_\textup{pf}=(x^\top,z^\top) Q_{\textup{E}} (x^\top,z^\top)^\top$. 
The weightings are 
$Q=\text{diag}(9\cdot 10^6, 9\cdot 10^6, 6, 1\cdot 10^2)$, 
$R=\text{diag}(6,6,6,6)$,
$Q_\textup{E}=\text{diag}(0,0,0,0,0,0,1\cdot 10^2,0)$.
The prediction horizon spans $T=\SI{150}{ms}$, while the sampling time is $T_\text{s}=\SI{10}{ms}$.

The state constraints are $\mathcal X=
[-170^\circ, 170^\circ ]
\times[-120^\circ, 120^\circ]
\times[-120^\circ, 120^\circ]
\times[-0.04\text{\,rad\,}\text{s}^{-1}, 0.04\text{\,rad\,}\text{s}^{-1} ]
\times[-1\text{\,rad\,}\text{s}^{-1},1\text{\,rad\,}\text{s}^{-1}]
\times[-0.05\text{\,rad\,}\text{s}^{-1},0.03\text{\,rad\,}\text{s}^{-1}]$. 
The virtual states in path following are constrained by $\mathcal Z=[-1,0]\times[\SI{0}{s^{-1}},\SI{1}{s^{-1}}]$.
 Furthermore, box constraints on the input and virtual inputs are present: $\mathcal U=[\SI{-13}{Nm},\SI{10}{Nm}]\times [\SI{-5}{Nm},\SI{5}{Nm}]\times [\SI{-5}{Nm},\SI{5}{Nm}]$, $\mathcal V=[\SI{-10}{s^{-2}},\SI{0.5}{s^{-2}}]$, while the contact force should satisfy $[\SI{0}{N},\SI{6}{N}]$.



\bibliography{ForceControl}

 \begin{IEEEbiography}[{\includegraphics[width=1in,height=1.25in,clip,keepaspectratio]{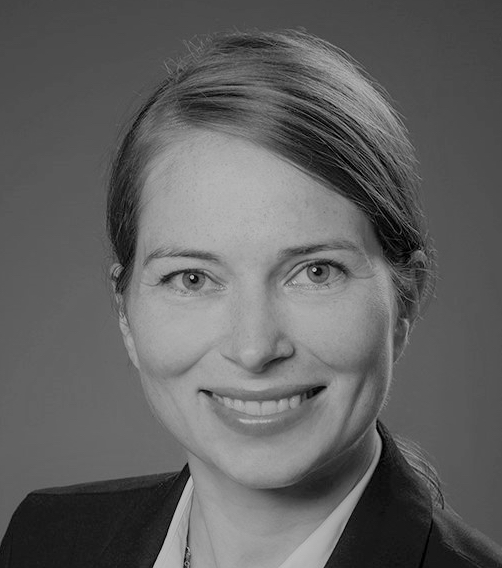}}]{Janine Matschek}
 obtained her bachelor's and master's degrees in systems engineering and engineering cybernetics from Otto von Guericke University Magdeburg, Germany, in 2010 and 2013, respectively. In 2021, she completed her Ph.D. in the field of model predictive control and machine learning at the University Magdeburg, Germany. She joined the Technical University of Darmstadt, Darmstadt, Germany, as a postdoctoral researcher in 2022.  Her research interests include machine learning for predictive control and control in medical and industrial robotic applications. Dr. Matschek received the award for the best dissertation of the faculty of electrical engineering at the Otto von Guericke University Magdeburg, Germany, in 2021.

\end{IEEEbiography}

\begin{IEEEbiography}[{\includegraphics[width=1in,height=1.25in,clip,keepaspectratio]{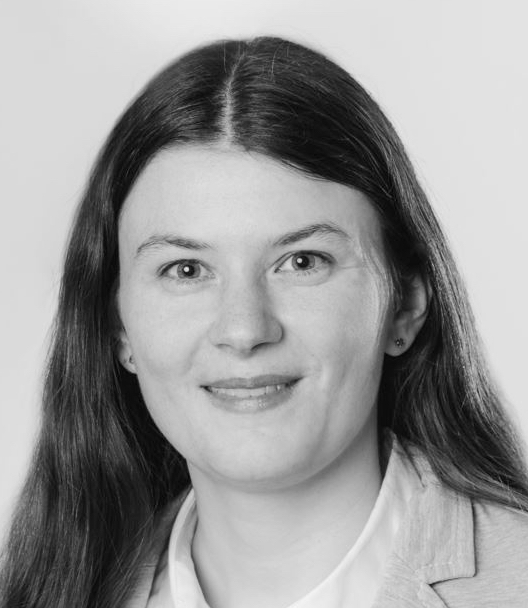}}]{Johanna Bethge} studied systems engineering and engineering cybernetics at Otto von Guericke University Magdeburg in Germany and Tampere University of Technology in Finland. She received her bachelor's and master's degrees in 2013 and 2016, respectively, from Otto von Guericke University Magdeburg. Since 2016, she has been a research assistant at the same university and is currently pursuing her Ph.D. Her research interests include learning-supported model predictive control and safety guarantees under uncertainty for multi-mode systems, such as autonomous vehicles. Johanna Bethge is a fellow of the Research Training Group for Mathematical Complexity Reduction (GRK 2297) of the German Research Foundation (DFG).

%
%
\end{IEEEbiography}

\begin{IEEEbiography}[{\includegraphics[width=1in,height=1.25in,clip,keepaspectratio]{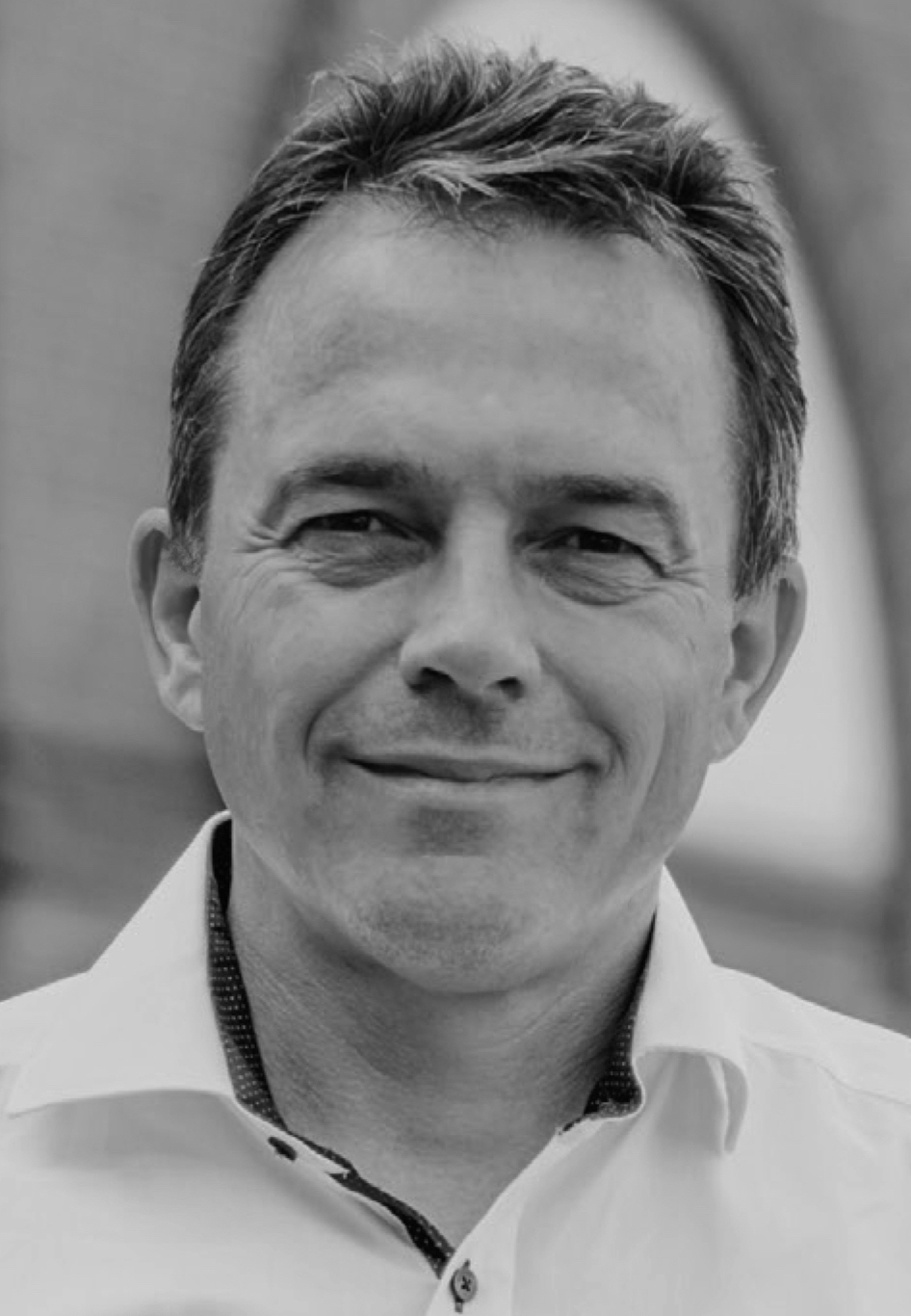}}]{Rolf Findeisen} earned a Diploma degree in Engineering Cybernetics from the University of Stuttgart, an M.S. degree from the University of Wisconsin-Madison in 1997, and his Ph.D. degree from the University of Stuttgart in 2005.

From 2007 to 2021, Rolf served as a full professor and led the Systems Theory and Automatic Control Laboratory at the Otto von Guericke University Magdeburg. Since August 2021, he has been heading the Control and Cyber-Physical Systems Laboratory at the Technical University of Darmstadt. Rolf and his group have published extensively in international journals, and he has served as an editor and associate editor for various prestigious publications, including IEEE Control Systems Magazine, IEEE Transactions on Networked Systems, J. Optimal Control Applications, and Methods and Processes. He also served as the IPC Co-Chair of the IFAC World Congress 2020.

Rolf's research focuses on the interplay between control and machine learning, autonomous systems, predictive control, cyber-physical systems, uncertainty, and robustness. His group's work has a wide range of applications, including robotics, autonomous driving, mechatronics, biotechnology, systems biology, battery and energy systems.

%
%
\end{IEEEbiography}

\end{document}